\documentclass[journal]{IEEEtran}
\usepackage{amsmath, amssymb}
\usepackage{algorithmic}
\usepackage{graphicx}
\usepackage{rotating}
\usepackage{comment}
\usepackage{amsmath}
\usepackage{subfig}
\usepackage{graphicx}
\usepackage{booktabs}
\usepackage{multirow}
\usepackage{colortbl}
\usepackage{hyperref}
\usepackage{algorithmic}
\usepackage{graphicx}
\usepackage{rotating}
\usepackage{comment}
\usepackage{amsmath}
\usepackage{subfig}
\usepackage[utf8]{inputenc}
\usepackage{amssymb}
\usepackage{amsfonts,amssymb,amsmath,amsgen,amsopn,amsbsy,theorem,graphicx,epsfig}
\usepackage{amscd,bezier,latexsym,mathrsfs,enumerate}
\usepackage[utf8]{inputenc}
\usepackage[english]{babel}
\usepackage{tabularx, booktabs}
\usepackage{cite}
\usepackage[table]{xcolor}
\usepackage[normalem]{ulem}
\usepackage{hyperref}
\def\subparagraph{} % because IEEE classes don't define this, but titlesec assumes it's present
\usepackage[compact]{titlesec} 
\titlespacing*{\section}{0pt}{2pt}{2pt} % {left}{before}{after}
\titlespacing*{\subsection}{0pt}{2pt}{2pt} % {left}{before}{after}
\titlespacing*{\subsubsection}{8pt}{1pt}{2pt} % Adjust {left}{before}{after} as needed
\usepackage[font=small, skip=2pt]{caption}
\usepackage[skip=1pt]{subcaption}    % Reduce vertical gap between subfigures and captions

\usepackage[nodisplayskipstretch]{setspace}
\hypersetup{
    colorlinks=true,
    linkcolor=magenta,      % internal links
    filecolor=magenta,      % links to local files
    urlcolor=magenta        % external links
}

\IEEEoverridecommandlockouts                              % This command is only needed if 
\usepackage{eso-pic}
\newcommand\AtPageUpperMyright[1]{\AtPageUpperLeft{%
 \put(\LenToUnit{0.5\paperwidth},\LenToUnit{-1.5cm}){%
     \parbox{0.5\textwidth}{\raggedright\fontsize{11}{11}\selectfont #1}}%
 }}%
\newcommand{\conf}[1]{%
\AddToShipoutPictureBG*{%
\AtPageUpperMyright{#1}
}
}
\title{\LARGE \bf
AFRDA: Attentive Feature Refinement for Domain Adaptive Semantic Segmentation
}

\conf{\hspace*{-9cm}\textcolor{gray}{This paper has been accepted for publication in \textbf{IEEE Robotics and Automation Letters} \textcolor{blue}{\href{https://ieeexplore.ieee.org/xpl/RecentIssue.jsp?punumber=7083369}{(RA-L)}}, July 2025.}}

\author{Md. Al-Masrur Khan, Durgakant Pushp, and Lantao Liu% <-this % stops a space
\thanks{%Md. Al-Masrur Khan, Durgakant Pushp, and Lantao Liu 
Authors are with the Luddy School of Informatics, Computing, and Engineering, Indiana University, Bloomington, IN 47408 USA (e-mail: \texttt{\{khanmdal, dpushp, lantao\}@iu.edu}).}%
\thanks {This work was supported by the National Science Foundation under Grant No. 2047169.}
}
\begin{document}

\maketitle

\thispagestyle{empty}
\pagestyle{empty}

%%%%%%%%%%%%%%%%%%%%%%%%%%%%%%%%%%%%%%%%%%%%%%%%%%%%%%%%%%%%%%%%%%%%%%%%%%%%%%%%
\begin{abstract}
In Unsupervised Domain Adaptive Semantic Segmentation (UDA-SS), a model is trained on labeled source domain data (e.g., synthetic images) and adapted to an unlabeled target domain (e.g., real-world images) without access to target annotations. Existing UDA-SS methods often struggle to balance fine-grained local details with global contextual information, leading to segmentation errors in complex regions. To address this, we introduce the Adaptive Feature Refinement (AFR) module, which enhances segmentation accuracy by refining high-resolution features using semantic priors from low-resolution logits. AFR also integrates high-frequency components, which capture fine-grained structures and provide crucial boundary information, improving object delineation. Additionally, AFR adaptively balances local and global information through uncertainty-driven attention, reducing misclassifications. Its lightweight design allows seamless integration into HRDA-based UDA methods, leading to state-of-the-art segmentation performance. Our approach improves existing UDA-SS methods by 1.05$\%$ mIoU on GTA V $\rightarrow$  Cityscapes and 1.04$\%$ mIoU on Synthia $\rightarrow$  Cityscapes. The implementation of our framework is available at: \href{https://github.com/Masrur02/AFRDA}{https://github.com/Masrur02/AFRDA} 

\end{abstract}

%%%%%%%%%%%%%%%%%%%%%%%%%%%%%%%%%%%%%%%%%%%%%%%%%%%%%%%%%%%%%%%%%%%%%%%%%%%%%%%%
\section{INTRODUCTION}

% 1st paragraph
Unsupervised domain adaptive semantic segmentation (UDA-SS) aims to train a model on a labeled source domain (e.g., synthetic data) and deploy it on an unlabeled target domain (e.g., real-world data) under significant appearance and distribution shifts \cite{tsai2018learning}. This is crucial for robotic vision, where models trained on curated or simulated datasets must generalize to new environments without additional annotation. UDA-SS methods are broadly categorized into adversarial learning-based \cite{gong2021dlow} and self-training (ST) approaches \cite{hoyer2022daformer}. ST methods have shown superior performance by addressing class-level misalignment and improving training stability \cite{chen2023ida}. Using a teacher-student framework, where the teacher generates pseudo-labels to supervise the student, ST offers a more robust solution. With DAFormer \cite{hoyer2022daformer}, UDA-SS saw a breakthrough via transformer-based self-supervised learning. This was further extended in HRDA \cite{hoyer2022hrda}, which integrates low and high-resolution features and now serves as the backbone for most recent UDA-SS methods.

While many existing approaches built on top of HRDA incorporate spatial and contextual details and leverage contrastive learning to enhance segmentation performance, they often overlook the critical issue of effectively integrating global context with fine-grained local details. High-resolution features capture fine-grained structures but lack broader contextual and class-level semantics, while low-resolution logits provide strong global priors and semantic consistency. However, most methods ignore incorporating these low-resolution logits into high-resolution features, leading to segmentation outputs that are spatially precise but contextually inconsistent. Furthermore, boundary information is often not effectively utilized, resulting in weak or imprecise delineation of object boundaries.

To address this problem, we propose enhancing HRDA-based UDA methods by incorporating boundary information along with global and local details. Therefore, we propose an Adaptive Feature Refinement (AFR) module for integration with HRDA-based methods. The AFR module refines high-resolution features by leveraging global information from low-resolution logits, enriching contextual details while simultaneously capturing fine-grained local spatial details from raw multi-scale high-resolution features. Additionally, AFR incorporates high-frequency components extracted from both high-resolution features and low-resolution logits, enabling enhanced edge consistency. It ensures robust boundary prediction by dynamically suppressing noisy predictions through uncertainty-driven attention and adaptively balancing local and global information using dual attention maps.

While inspired by prior attention and boundary refinement methods, our AFR module introduces an integration tailored for domain adaptation. Although its individual components build upon existing works, their coordinated integration into a lightweight, plug-and-play dual-attention module is, to the best of our knowledge, unique in the UDA literature. Unlike existing methods that operate solely in the feature space, AFR is the first to incorporate semantic logits directly into the refinement process. It places semantic logits, uncertainty, and high-frequency components at the core of its design, rather than treating these elements as auxiliary cues, enabling semantically consistent and boundary-aware high-resolution refinement without an explicit boundary head. AFR is designed to keep the existing training pipeline unchanged and instead focus on modular refinement at the feature level. This enables AFR to act as a plug-and-play component within HRDA-based UDA frameworks without disrupting training dynamics. This modularity is particularly important for real-world robotic deployment, where maintaining training stability and simplicity is crucial.
 % Last paragraph
We validate our approach on \textit{five} challenging datasets spanning both urban and off-road environments. In the standard GTA V~\cite{richter2016playing} $\rightarrow$ Cityscapes~\cite{cordts2016cityscapes} and SYNTHIA~\cite{ros2016synthia} $\rightarrow$ Cityscapes setups, AFRDA outperforms prior methods with improved segmentation quality and finer detail preservation. More importantly, we extend evaluation to off-road adaptation from RUGD~\cite{8968283} to our in-house forest dataset (MESH), where AFRDA maintains strong performance in unstructured terrains with vegetation, uneven ground, and natural obstacles—an area often neglected in UDA-SS research. To demonstrate real-world applicability, we deploy AFRDA on a robot navigating an outdoor environment, where our model enables accurate, stable perception of traversable ground and obstacles, supporting reliable autonomous navigation.

\section{RELATED WORKS}

\subsection{Unsupervised Domain Adaptation}
Unsupervised Domain Adaptation (UDA) bridges the domain gap between labeled source and unlabeled target domains and adapts semantic segmentation models across diverse environments. Numerous strategies have been proposed to improve UDA performance. Among them, methods based on statistical distance functions utilize 
%correlation alignment \cite{sun2016deep}
entropy minimization \cite{vu2019advent}, or Wasserstein distance \cite{lee2019sliced} to mitigate domain discrepancy. Another line of UDA works leverages the adversarial learning paradigm \cite{gong2021dlow}, where a learned domain discriminator globally aligns features from both domains. As adversarial learning ignores class-level alignment, it fails to eliminate class-level shifts and suffers from the negative transfer problem. %cite{zhang2022survey}. 
Though some methods utilize class-level features, the lack of target labels often leads to weak performance and unstable training. Self-training-based approaches \cite{hoyer2022daformer} have emerged as a promising solution, using a teacher-student framework where the teacher model generates pseudo labels for the target domain. However, noisy pseudo-labels frequently reduce the effectiveness and performance of these methods. Recent methods explore multi-crop consistency \cite{hoyer2022hrda}, contextual clues \cite{hoyer2023mic}, contrastive learning \cite{xiang2024pseudolabel}, and auxiliary refinement networks \cite{zhao2024unsupervised} to create reliable pseudo-labels and improve UDA.

\subsection{Feature Refinement in Semantic Segmentation}
%Semantic segmentation models often suffer from spatial and contextual inconsistencies, leading to misclassifications, especially around object boundaries. To address this, researchers have explored feature refinement techniques to enhance both spatial accuracy and contextual understanding. Among the feature refinement techniques, one widely adopted strategy is multi-scale feature refinement. Methods like Feature Pyramid Networks (FPN) \cite{lin2017feature}, Atrous Spatial Pyramid Pooling (ASPP) \cite{chen2017deeplab}, and HRNet \cite{wang2020deep} effectively capture both fine-grained details and global contextual information by integrating features from multiple resolutions. Another powerful refinement strategy leverages attention mechanisms \cite{woo2018cbam}, \cite{hu2018squeeze} to focus on the most relevant features.
Semantic segmentation models often suffer from spatial and contextual inconsistencies, leading to misclassifications, especially around object boundaries. To address this, researchers have explored feature refinement techniques to improve both spatial accuracy and contextual understanding. A widely adopted strategy is multi-scale feature refinement. Methods like Feature Pyramid Networks (FPN) \cite{lin2017feature}, Atrous Spatial Pyramid Pooling (ASPP) \cite{chen2017deeplab}, and HRNet \cite{wang2020deep} capture fine-grained details and global context by integrating features from multiple resolutions. Another effective refinement strategy leverages attention mechanisms \cite{woo2018cbam}, \cite{hu2018squeeze} to focus on the most relevant features.

While attention modules like CBAM\cite{woo2018cbam} and SE-Net\cite{hu2018squeeze} refine features by dynamically prioritizing important spatial and channel information, enhancing segmentation accuracy in complex scenes, they do not incorporate semantic logits or uncertainty, which are central to our AFR design. Moreover, edge and boundary-guided refinement has also been widely explored to address mispredictions around object edges. Techniques such as Gated-SCNN \cite{takikawa2019gated} and STDC-Seg \cite{fan2021rethinking} incorporate explicit boundary supervision by employing additional boundary heads to improve object delineation, while some methods design boundary-aware loss functions to enhance feature quality near object edges. In contrast, our AFR module captures boundary information implicitly through high-frequency residuals, eliminating the need for extra labels or specialized boundary heads. Moreover, recent uncertainty-aware domain adaptation methods such as UPA \cite{chen2024uncertainty} employ pixel-level uncertainty primarily for filtering unreliable pseudo-labels in a source-free setting, rather than integrating uncertainty into attention-guided multi-resolution feature refinement as done in our AFR module. Our AFR module uniquely combines global semantic logits, boundary-enhancing high-frequency signals, and dual uncertainty maps for adaptive high-resolution feature refinement, which is structurally and functionally distinct from these prior approaches.

%Recently, UDA methods are also exploring feature refinement techniques due to their capability to capture fine-grained spatial details and global contextual information, which are essential for addressing the domain gap and improving segmentation performance in target domains.
%\vspace{-12pt}
\subsection{Visual Navigation using Semantic Segmentation}
%Semantic segmentation is important for visual navigation as it classifies terrain as traversable, non-traversable, forbidden, etc. Traditional RGB-based segmentation models can generate 2D maps, distinguishing navigable and non-navigable areas \cite{guan2022ga}. However, these methods cannot often sense surface slope and elevation height, especially for off-road navigation, and struggle to differentiate between terrain classes that are technically navigable but may not be optimal for traversal (e.g., small vs. tall grasses). They may even get stuck in local minima. Incorporating geometric information from LiDAR point clouds along with segmentation maps can address these issues by producing a 2.5D occupancy grid map \cite{guan2021tns}, \cite{maturana2018real} 
%allowing motion planners to create semantically aware trajectories. More recently, integrating proprioceptive sensing (i.e., linear and angular velocity, force, torque) with geometric and visual cues has become widely popular due to its capability of creating robust 2D and 3D cost maps \cite{sivaprakasam2024salon} %, \cite{jiao2024real} 
%that consider surface bumpiness and roughness, helping to enable context-aware navigation. In our proposed work, we focus on leveraging semantic segmentation to enhance visual navigation in complex environments.
Semantic segmentation is important for visual navigation as it classifies terrain as traversable, non-traversable, forbidden, etc. Traditional RGB-based segmentation models can generate 2D maps, distinguishing navigable and non-navigable areas \cite{guan2022ga}. However, these methods often cannot sense surface slope and elevation height—especially for off-road navigation—and struggle to differentiate between terrain classes that are technically navigable but may not be optimal for traversal (e.g., small vs. tall grasses). They may even get stuck in local minima. Incorporating geometric information from LiDAR point clouds along with segmentation maps can address these issues by producing a 2.5D occupancy grid map \cite{guan2021tns}, allowing motion planners to create semantically aware trajectories. More recently, integrating proprioceptive sensing (i.e., linear and angular velocity, force, torque) with geometric and visual cues has become widely popular due to its capability of creating robust 2D and 3D cost maps \cite{sivaprakasam2024salon}, considering surface bumpiness and roughness to enable context-aware navigation. In our proposed work, we focus on leveraging semantic segmentation to enhance visual navigation in complex environments.

\section{METHODOLOGY}
%First, in Sect. \ref{Pre}, we provide the preliminaries of UDA methods and an overview of our proposed framework, AFRDA. Then, in Sect. \ref{AFR}, we introduce our designed component, the Attentive Feature Refinement (AFR) module, which enhances segmentation accuracy and boundary stabilization. Finally, in Sect. \ref{V}, we discuss the visual planner used to deploy our framework in a robotic vehicle.
First, in Sect. \ref{Pre}, we provide the preliminaries of UDA methods and an overview of our proposed framework, AFRDA. Then, in Sect. \ref{AFR}, we introduce the Attentive Feature Refinement (AFR) module, which enhances segmentation accuracy and boundary stabilization. Finally, in Sect. \ref{V}, we discuss the visual planner used to deploy our framework in a robotic vehicle.

\subsection{Preliminaries of UDA-SS}\label{Pre} 

\begin{figure*}[t] \vspace{-5pt}
    \centering
   \includegraphics[width=1\linewidth]{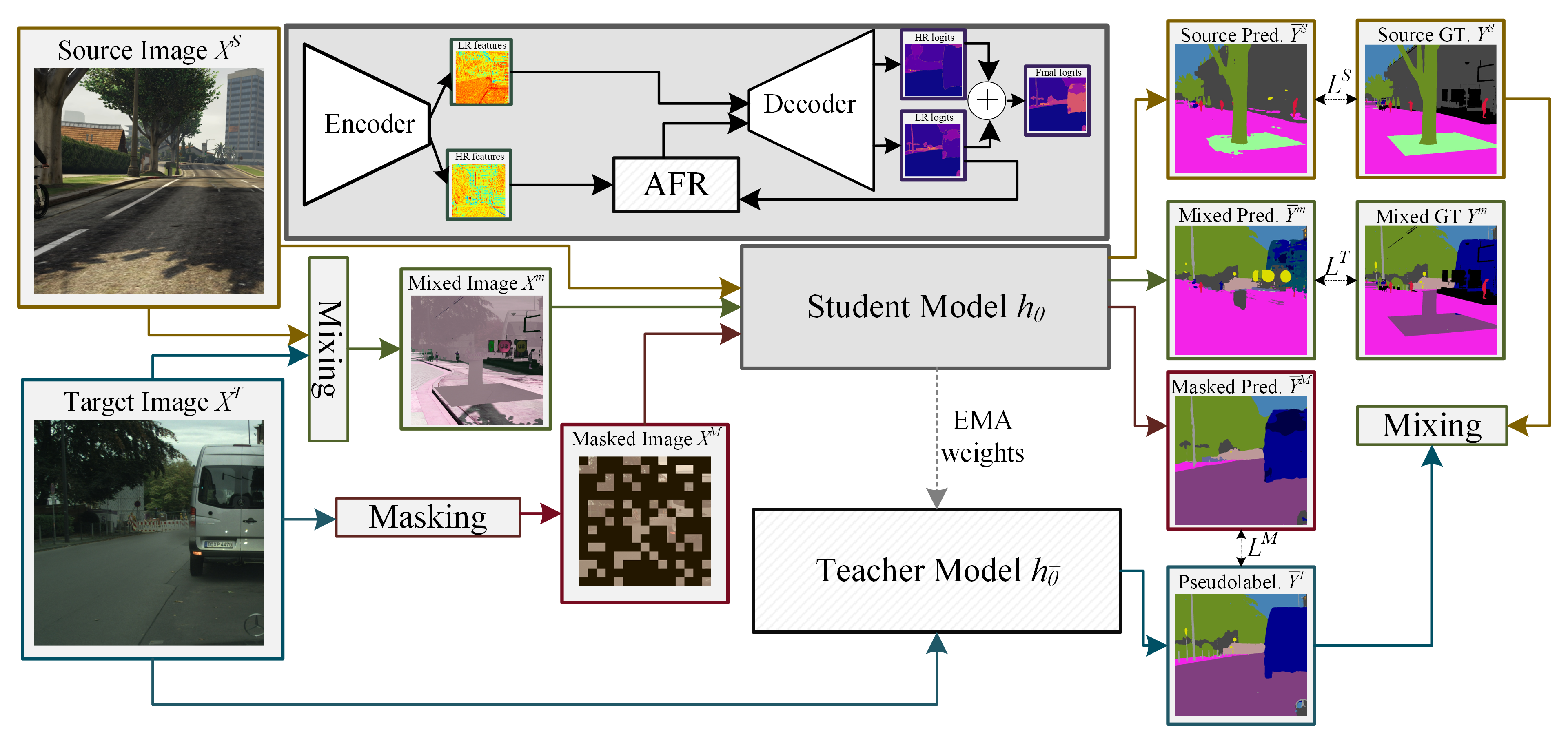} % Replace 'example-image' with your immixedage filename

    \vspace{-5pt}
    \caption{\small A brief illustration of our proposed Attentive Feature Refinement for Domain Adaptation (AFRDA) model based on a teacher-student architecture. Given the labeled source data ${X^S, Y^S}$, we use the network $h_{\theta}$ to compute the segmentation prediction $\bar{Y}^S$, supervised by the source loss $L^S$. The AFR module is incorporated in the student model, and the gray block on top shows its internal mechanism. The teacher model $\bar{h}_{\theta}$ generates pseudo-labels $\bar{Y}^T$ for target images $X^T$ without backpropagation. Then, mixed images $X^m$ and labels $y^m$ are created by mixing source and target data, and the student predicts $\bar{Y}^m$ supervised by loss $L^T$. Finally, we mask target images $X^T$ to obtain $X^M$ and use the student to predict $\bar{Y}^M$ using $\bar{Y}^T$ as ground truth.\vspace{-10pt}}
    \label{m1} 
\end{figure*}
In UDA, we consider a source domain \textit{S} with $N_s$ labeled images ($X^S$=$\{x_i^S\}_{i=1}^{N_s}$, $Y^S$=$\{y_i^S\}_{i=1}^{N_s}$) and a target domain with $N_t$ raw images $X^T$=$\{x_i^T\}_{i=1}^{N_t}$ without any access to the ground truth $Y^T$. A neural network \textit{h} parameterized by $\theta$ is trained on the source domain images $X^S$ and is adapted on the target domain images $X^T$. However, training only on the source domain data with a categorical cross-entropy loss,
\begin{equation}
L_i^S = - \sum_{j=1}^{H \times W} \sum_{c=1}^{C} y_i^S \log h_{\theta}(x_i^S)
\end{equation}
where \textit{C} denotes the number of semantic classes, does not generalize well to target domain images,  resulting in poor target domain predictions. To address this domain shift, recent UDA methods like DAFormer \cite{hoyer2022daformer}, HRDA \cite{hoyer2022hrda} adopt the self-training strategy. These methods utilize a teacher model $h_{\bar{\theta}}$ to generate pseudo-labels $\bar{Y}^{T} = \textit{argmax} \; h_{\bar{\theta}}(X^{T})$ without the backpropagation of any gradients. The weights of the teacher model $h_{\bar{\theta}}$ get updated using an Exponential Moving Average (EMA) based on the weights of the student model $h_\theta$, after each training iteration \textit{t}
\vspace{-2pt}
\begin{equation}
\bar{\theta}_{t+1} \leftarrow \alpha \bar{\theta}_t + (1 - \alpha) \theta_t.
\end{equation}

The generated pseudo-labels can be noisy, so a quality estimate $q^T$ is produced based on the proportion of pixels whose maximum softmax probability exceeds a threshold $\tau$. Mathematically,
\begin{equation}
q^T_i = \frac{\sum_{j=1}^{H \times W}  \left[ \max_{c} \bar{h}_{\theta}(x^T_i) > \tau \right]}{H \cdot W}.
\end{equation}
Later, the student model $h_\theta$ is trained again on the target using the pseudo labels and their quality estimates to optimize the target domain loss
\begin{equation}
L_i^T = - \sum_{j=1}^{H \times W} \sum_{c=1}^{C} \bar{q}_i^T \bar{y}_i^T \log h_{\theta}(x_i^T).
\end{equation}
However, recent methods like DAFormer incorporate ClassMix \cite{olsson2021classmix} and other mixing techniques \cite{chen2023ida} to generate mixed pairs ($X^m$=$\{x_i^m\}_{i=1}^{N_m}$, $Y^m$=$\{y_i^m\}_{i=1}^{N_m}$) by blending source and target domain images, labels, and pseudo-labels. Instead of directly training on the pseudo-labels, the student model is trained on these mixed images, mitigating confirmation bias, enhancing robustness, and improving adaptation across domains. Hence, the target domain loss becomes
\begin{equation}
L_i^T = - \sum_{j=1}^{H \times W} \sum_{c=1}^{C} y_i^m \log h_{\theta}(x_i^m).
\end{equation}

This approach improves adaptation to the target domain. However, it still faces challenges in accurately segmenting small objects and preserving fine details due to the use of only low-resolution input, which limits the ability to capture fine-grained structures. HRDA addressed this by introducing a multi-resolution framework that combines a large low-resolution (LR) context crop for long-range dependencies and a small high-resolution (HR) detail crop for fine segmentation.

Within the HRDA framework, we employ an Adaptive Feature Refinement (AFR) module to enhance the feature representation for different classes (\textcolor{blue}{see Fig.~\ref{m1}}). The AFR module refines high-resolution features by leveraging global information from low-resolution logits, enriching contextual details while capturing fine-grained spatial details from raw multi-scale high-resolution features. Furthermore, to leverage contextual clues, random patches are masked out from the target images, generating a set of masked target images $X^M = \{x_i^M\}_{i=1}^{N_t}$ in UDA methods~\cite{hoyer2023mic}. Finally, the student model is trained on the masked images supervised by target domain pseudo labels $\bar{y}_i^T$ to optimize the masked loss.

\begin{equation}
L_i^M = - \sum_{j=1}^{H \times W} \sum_{c=1}^{C} \bar{q}_i^T \bar{y}_i^T \log h_{\theta}(x_i^M).
\end{equation}
Hence combining the preliminaries of UDA and our proposed AFR module the overall adaptation objective of our framework becomes
\begin{equation}
\min_\theta\mathcal{L}= L^S  + L^T  + L^M.
\end{equation}
\begin{figure*}[t]
    \centering
    \includegraphics[width=1\linewidth]{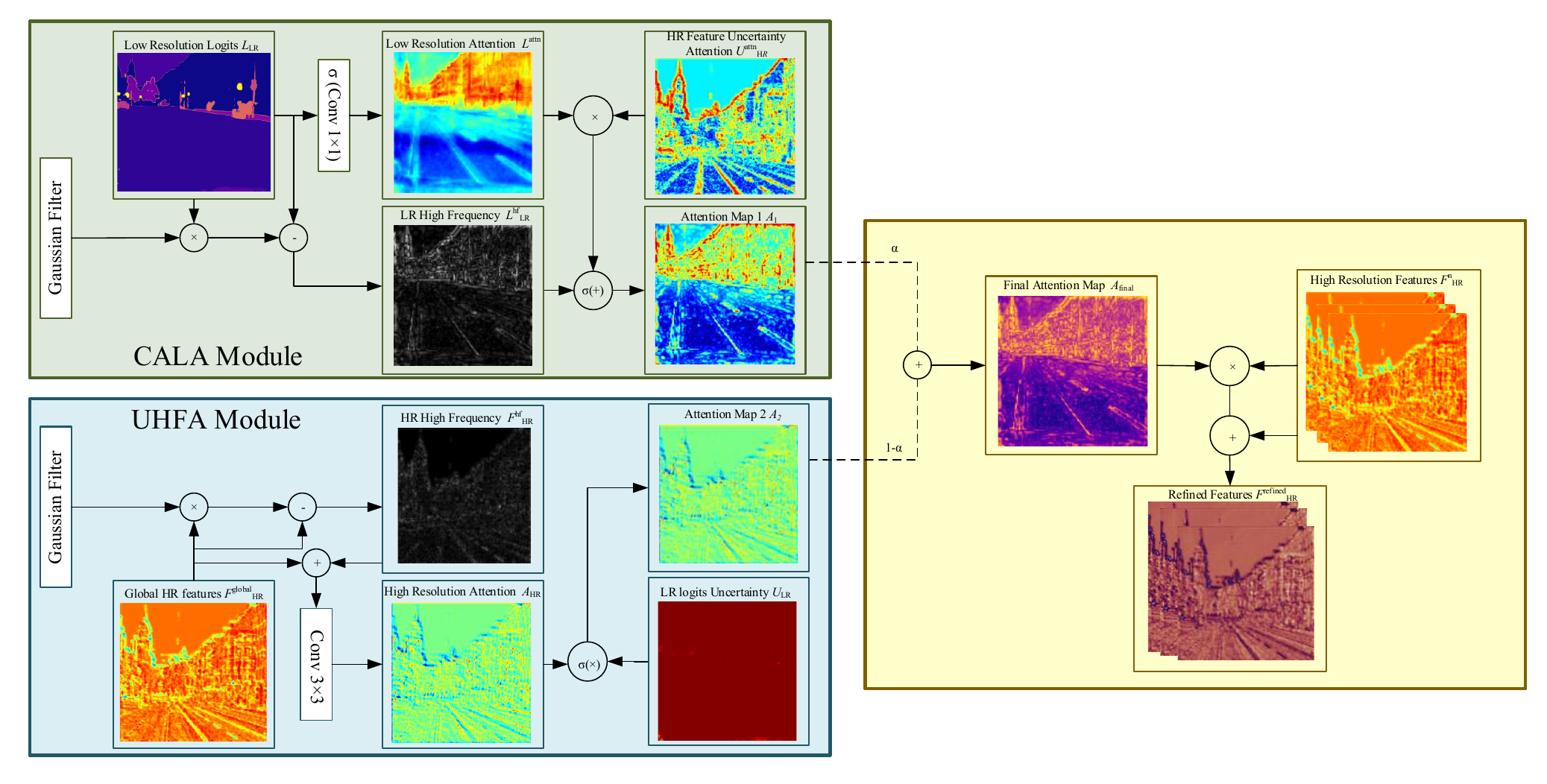} % Replace 'example-image' with your immixedage filename
    \caption{\small Framework overview of the Attentive Feature Refinement Module. AFR consists of two components: the CALA and UHFA modules. Both generate attention maps, which are combined using a learnable parameter $\alpha$ to produce the final attention map $A_\text{final}$, used to refine the high-resolution feature maps $F_{\text{HR}}^{n}$.}
    \vspace{-17pt}
    \label{f2}
\end{figure*}
\subsection{Attentive Feature Refinement} \label{AFR}
%To recognize an object, a model needs to utilize both high-resolution and low-resolution representations. High-resolution features can capture fine-grained spatial details, and the low-resolution features can provide global context. Many UDA architectures can integrate both HR and LR features in their frameworks to balance local accuracy with large-scale contextual reasoning. While feature fusion can enhance the representation of both spatial and contextual information, it ignores the class-aware information from low-resolution logits that can provide a structural refinement signal for high-resolution features. 
To recognize an object, a model needs to utilize both high-resolution and low-resolution representations. High-resolution features capture fine-grained spatial details, and low-resolution features provide global context. Many UDA architectures integrate both HR and LR features to balance local accuracy with large-scale contextual reasoning. While feature fusion can enhance the representation of both spatial and contextual information, it ignores class-aware information from low-resolution logits that can provide a structural refinement signal for high-resolution features.

Unlike encoder features, which are optimized for internal representation and may contain redundant or noisy activations, low-resolution logits are supervised outputs that explicitly encode class-level semantics. As a result, they are more interpretable, semantically aligned, and closer to the final prediction output. Additionally, softmax-normalized logits enable confidence estimation, which we exploit through uncertainty-based refinement. This makes logits more suitable for guiding high-resolution features, ensuring better class-level consistency and computational efficiency in the refinement process.
%Therefore, we propose to leverage the global class distributions and uncertainty from low-resolution logits to guide the refinement of high-resolution features, ensuring that they incorporate both spatial details and semantic consistency. In order to facilitate the refinement of HR features guided by LR logits, we introduce the Attentive Feature Refinement (AFR) module, which can be easily integrated into various existing HRDA-based UDA methods. 
Therefore, we propose to leverage the global class distributions and uncertainty from low-resolution logits to guide the refinement of high-resolution features, ensuring they incorporate both spatial details and semantic consistency. To facilitate the refinement of HR features guided by LR logits, we introduce the Attentive Feature Refinement (AFR) module, which can be easily integrated into various existing HRDA-based UDA methods. Although AFR does not directly interact with pseudo-label filtering, adaptive weighting, or mask correction, its uncertainty-aware refinement indirectly complements these processes by suppressing noisy features and stabilizing predictions. This results in a stronger student model and, over time, higher-quality, cleaner pseudo-labels through EMA-based teacher updates. The feature refinement process AFR is illustrated in Fig. \ref{f2} and described below.

AFR preserves spatial details and integrates semantic consistency by incorporating two complementary attention mechanisms: Class-Aware Logits-Based Attention (CALA) and Uncertainty-Suppressed HR Feature Attention (UHFA). 
\subsubsection*{\textbf{Class-Aware Logits-Based Attention}} The Class-Aware Logits-based Attention extracts class-aware information from low-resolution logits while integrating uncertainty maps from the HR features. Uncertainty maps are estimated from softmax probabilities due to their computational simplicity and direct availability from the segmentation output, unlike entropy- or calibration-based methods. The module first takes the LR logits $L_{\text{LR}} \in \mathbb{R}^{B \times C \times H \times W}$ as input and passes through a 1$\times$1 convolution to reduce the multi-channel representation into a single channel attention map 
\begin{equation}
L^{\text{attn}}=\sigma (conv^{1\times1}(L_{\text{LR}})),
\end{equation}
where $\sigma$ denotes the sigmoid activation function. In the later part of the CALA module, the uncertainty maps of the HR features $U_{\text{HR}} \in \mathbb{R}^{B \times 1 \times H \times W}$ are also passed through a sigmoid activation function
\begin{equation}
U_{\text{HR}}^{\text{attn}} = \sigma(U_{\text{HR}}).
\end{equation}
These two attention maps $L^{\text{attn}}$ and $U_{\text{HR}}^{\text{attn}}$ are then combined through an element-wise multiplication for attaining a modulated attention map,
\begin{equation}
A_{\text{logits}} = L^{\text{attn}} \odot U_{\text{HR}}^{\text{attn}}
\end{equation}
that ensures low-resolution global information is emphasized in uncertain HR regions while maintaining high-resolution spatial precision where HR features are reliable. After obtaining the modulated attention map, the CALA module extracts the high-frequency component of the LR logits 
\begin{equation}
L_{\text{LR}}^{\text{hf}} = L_{\text{LR}} - G_{\gamma}^{2D} \otimes \label{eq1}
 L_{\text{LR}}
\end{equation}
using a Gaussian Smoothing filter $G_{\gamma}^{2D}$ and subtracting the smoothed version from the original logits. The Gaussian smoothing produces smooth, differentiable transitions that are well-suited for semantic segmentation. Unlike edge detectors like Sobel, which generate sharp, binary, and non-differentiable edges, Gaussian filtering captures soft boundary variations and preserves semantic continuity across overlapping class boundaries. Its differentiability supports gradient-based optimization, allowing seamless integration into end-to-end training without requiring an additional boundary prediction head. The Gaussian Smoothing filter $G_{\gamma}^{2D}$ is constructed by taking the outer product of two 1D Gaussian functions. Mathematically,
\begin{equation} 
G_{\gamma}^{2D}(i, j) = \frac{1}{2\pi \gamma^2} \exp\left(-\frac{i^2 + j^2}{2\gamma^2}\right),\, \text{for} \, i, j \in \left[ -\frac{k}{2}, \frac{k}{2} \right]
\end{equation}
where $\gamma$ is the standard deviation and \textit{k} is the kernel size. Finally, the logit-guided attention map is computed by integrating the modulated logits and the high-frequency components of raw LR logits
\begin{equation}
A_1 = \sigma(A_{\text{logits}} + L_{\text{LR}}^{\text{hf}})
\end{equation}
that incorporate the boundary information with global class distribution and uncertainty-based refinement.
\subsubsection*{\textbf{Uncertainty-Suppressed HR Feature Attention}}
The Uncertainty-Suppressed HR Feature Attention enhances the HR features by incorporating raw HR features and the uncertainty information from the LR logits. The UHFA module first takes the HR feature maps $F_{\text{HR}} = \{\, F_{\text{HR}}^{(n)} \,\}_{n=1}^{N}, \quad F_{\text{HR}}^{(n)} \in \mathbb{R}^{B \times C_n \times H_n \times W_n}$ as input and apply a global average pooling (GAP) operation across the channel dimension to compute a global feature representation $F_{\text{HR}}^{\text{global}} \in \mathbb{R}^{B \times 1 \times H \times W}$. Later the UHFA module extracts the high-frequency components from the $F_{\text{HR}}^{\text{global}}$ by using Gaussian Smoothing. Mathematically, 
\begin{equation}
F_{\text{HR}}^{\text{hf}} = F^{\text{global}}_{\text{HR}} - G_{\gamma}^{2D} \otimes
 F^{\text{global}}_{\text{HR}},
\end{equation}
where, $G_{\gamma}^{2D}$ is calculated by using the Eqn. \eqref{eq1}. After that, the globally pooled HR features $F_{\text{HR}}^{\text{global}}$ and its high-frequency counterpart $F_{\text{HR}}^{\text{hf}}$ are fused to emphasize boundary structure and then passed through spatial attention based on a $3\times3$ convolution attention layer
\begin{equation}
A_{\text{HR}} = \text{Conv}^{(3 \times 3)}(F_{\text{HR}}^{\text{global}} + F_{\text{HR}}^{\text{hf}})
\end{equation}
for selectively enhancing important spatial regions while suppressing less informative areas, ensuring that boundary structures and ambiguous class regions receive higher attention for improved feature discrimination. In the next step of the UHFA, the $A_{\text{HR}}$ and the exponential form of the uncertainty map from LR logits $U_{\text{LR}} \in \mathbb{R}^{B \times 1 \times H \times W}$ undergo an element-wise multiplication and are then followed by a sigmoid activation function to obtain the Uncertainty-Suppressed HR Feature Attention map
\begin{equation}
A2 = \sigma(A_{\text{HR}} \otimes \exp(- U_{\text{LR}})).
\end{equation}
This attention map prioritizes HR features in regions where LR logits are confident, ensuring that fine-grained spatial details are preserved in segmentation boundaries. Conversely, it suppresses the influence of HR features in uncertain LR regions to prevent overfitting to potentially unreliable high-resolution details that lack global class priors. After obtaining both the attention maps \textit{A1} and \textit{A2}, the AFR module combines them adaptively to generate the final attention map 
\begin{equation}
A_{\text{final}} = \alpha A1 + (1 - \alpha) A2
\end{equation}
where $\alpha$ is a learnable parameter. Finally, the AFR module computes the refined HR features by performing an element-wise multiplication between the raw HR features and the final attention map followed by a residual connection
\begin{equation}
F_{\text{HR}}^{\text{refined}} = (F_{\text{HR}} \odot A_{\text{final}} )+ F_{\text{HR}}.
\end{equation}
Therefore, the AFR module produces refined HR features that are spatially coherent, boundary-aware, and semantically consistent while also preserving the raw HR features to enhance robustness.
%\vspace{-7.9pt}
\subsection{Visual Planning}\label{V}
\vspace{-3pt}
%We integrate the proposed AFRDA module with our recently developed POVNav visual planner~\cite{pushp2023povnav} to enable efficient vision-based navigation. First, AFRDA generates a \textit{Navigability Image} $I^n_t$ by classifying pixels into navigable ($\Omega_N$) and non-navigable ($\Omega_{NN}$) regions, and POVNav applies the \textit{Visual Horizon} concept to ensure clear separation between them. The 3D goal is then projected onto the image border as the \textit{Peripheral Optic Goal} (POG), and a Pareto-optimal sub-goal, termed the \textit{Horizon Optic Goal} (HOG), is selected on the visual horizon by balancing deviation from the goal direction and forward progress. A collision-free visual path is planned from the robot’s position to the HOG, and two key features—\textit{proximity} ($\lambda$), measuring the relative distance to the horizon, and \textit{alignment} ($\phi$), quantifying path deviation—are extracted. The control error $e(t) = [\lambda, \phi] - [\lambda_0, 0]$ is minimized using a visual servoing scheme, where $\lambda$ regulates forward velocity and $\phi$ adjusts angular velocity. This allows the robot to navigate smoothly toward the goal while adapting to dynamic obstacles. 
We integrate the proposed AFRDA module with our recently developed POVNav visual planner~\cite{pushp2023povnav} to enable efficient vision-based navigation. AFRDA generates a \textit{Navigability Image} $I^n_t$ by classifying pixels into navigable ($\Omega_N$) and non-navigable ($\Omega_{NN}$) regions, while POVNav applies the \textit{Visual Horizon} concept to ensure clear separation. The 3D goal is projected onto the image border as the \textit{Peripheral Optic Goal} (POG), and a Pareto-optimal sub-goal, termed the \textit{Horizon Optic Goal} (HOG), is selected by balancing deviation from goal direction and forward progress. A collision-free visual path is planned to the HOG, and two features—\textit{proximity} ($\lambda$), measuring distance to the horizon, and \textit{alignment} ($\phi$), quantifying path deviation—are extracted. The control error $e(t) = [\lambda, \phi] - [\lambda_0, 0]$ is minimized via visual servoing, where $\lambda$ regulates forward and $\phi$ adjusts angular velocity. This enables smooth, adaptive navigation toward the goal.
\section{EXPERIMENTS}
\subsection{Evaluation Setup}
\subsubsection{\textbf{Datasets}} 
%In this work, we use five datasets to evaluate our framework. Among the datasets, GTA V is a synthetic dataset collected from a game environment. The GTA V dataset consists of 24,966 high-resolution images with corresponding labels for 33 classes. Like GTA V, Synthia is also a simulation dataset with 9400 images. The Synthia dataset shares 16 semantic classes with the GTA V dataset. Another urban dataset used in our work is the Cityscapes dataset. The Cityscapes dataset has 2975 training and 500 validation images. Among the forest datasets, one is the RUGD dataset which is an off-road dataset consisting of 7453 images with 24 semantic classes and 8 different unique terrain types. Another forest dataset, the MESH dataset, was collected in front of our lab and consists of 4,415 training images and 827 validation images.
In this work, we use five datasets to evaluate our framework. Among the datasets, GTA V is a synthetic dataset collected from a game environment, consisting of 24,966 high-resolution images with labels for 33 classes. Like GTA V, Synthia is also a simulation dataset with 9400 images and shares 16 semantic classes with the GTA V dataset. Another urban dataset used in our work is the Cityscapes dataset, which has 2975 training and 500 validation images. Among the forest datasets, one is the RUGD dataset, an off-road dataset with 7453 images, 24 semantic classes, and 8 unique terrain types. The MESH dataset was collected in front of our lab and consists of 4,415 training images and 827 validation images.

\subsubsection{\textbf{Implementation Details}}
%We develop our AFRDA model on the top of the SOTA method MIC \cite{hoyer2023mic}. We use a batch size of 2 and a crop size of 952 due to limited GPU memory. We have evaluated our model on the validation set of Cityscapes and MESH datasets by considering the mean Intersection over Union (mIoU) for the city environment and qualitative results for forest adaptation as the MESH dataset has no ground truth.
We develop our AFRDA model on top of the SOTA method MIC \cite{hoyer2023mic}, using a batch size of 2 and a crop size of 952 due to limited GPU memory. We evaluate the model on the validation set of Cityscapes and MESH datasets, considering mean Intersection over Union (mIoU) for the city environment and qualitative results for forest adaptation, as the MESH dataset has no ground truth.
\subsection{Result Comparison}
We compare our proposed AFRDA with the baseline, MIC~\cite{hoyer2023mic}, along with other SOTA methods, both quantitatively and qualitatively. We conduct two domain adaptation tasks in an urban environment: \textbf{GTA V $\rightarrow$ Cityscapes} and \textbf{Synthia $\rightarrow$ Cityscapes}, as well as a forest adaptation task from \textbf{RUGD $\rightarrow$ MESH}. First, we show the quantitative results on GTA V $\rightarrow$ Cityscapes adaptation in Table \ref{gt}.

\begin{figure*}[h] \vspace{-5pt}{
    \centering
    \includegraphics[width=1\linewidth]{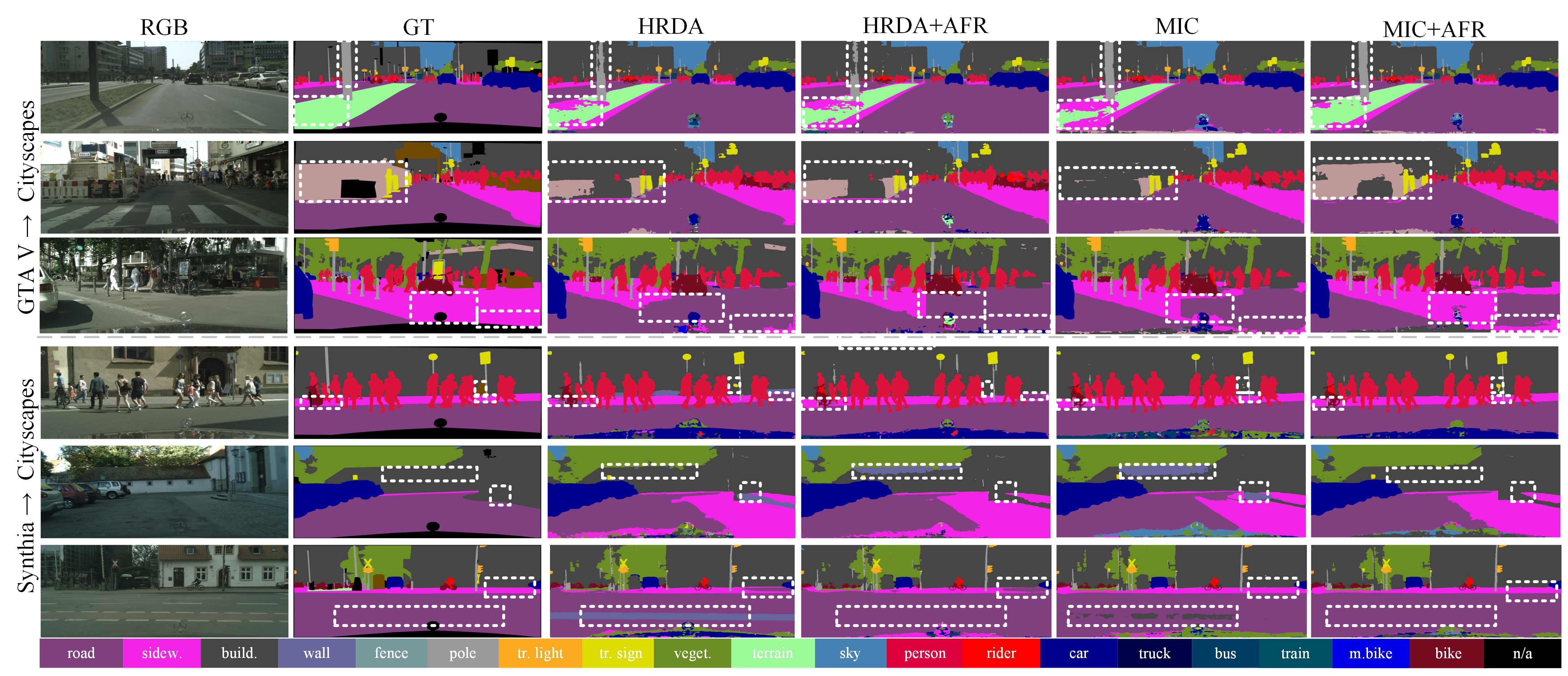} % Replace 'example-image' with your immixedage filename
    \caption{\small Qualitative Results for the adaptation of GTA V $\rightarrow$ Cityscapes and Synthia $\rightarrow$ Cityscapes.}
    
    \label{cityp} %\vspace{-25pt}
}
\end{figure*}

\begin{figure*}[h]
{    \centering
    \includegraphics[width=1\linewidth]{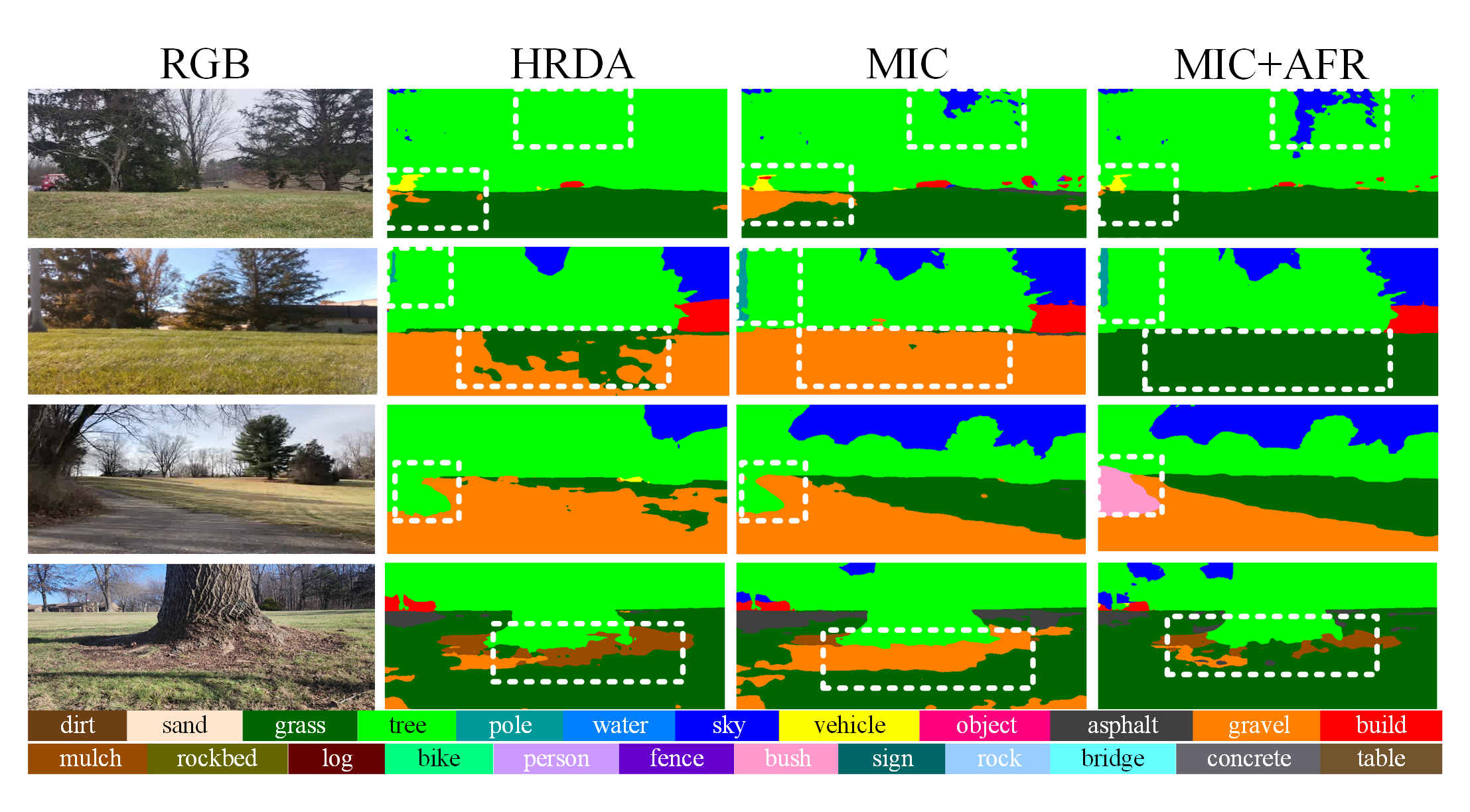} % Replace 'example-image' with your immixedage filename
   % \vspace{-10pt}
    \caption{\small Qualitative Results for the adaptation of RUGD $\rightarrow$ MESH. }
    \label{meshp} }
\end{figure*}

\begin{table*}[h]
    \centering
    \renewcommand{\arraystretch}{1.3} % Optional: Adjust row spacing
    \caption{Quantitive comparison of the adaptation from GTA V $\rightarrow$ Cityscapes dataset.  }
    \label{gt}
    \resizebox{\textwidth}{!}{ % Scale table to page width
    \begin{tabular}{p{2.2cm}|*{20}{>{\centering\arraybackslash}p{0.75cm}}} % Adjust column widths
        \toprule
        Method & Road & S.walk & Build. & Wall & Fence & Pole & Tr.Light & Sign & Veget. & Terrain & Sky & Person & Rider & Car & Truck & Bus & Train & M.bike & Bike & mIoU \\ \hline
        
        ADVENT \cite{vu2019advent} & 89.4 & 33.1 & 81.0 & 26.6 & 26.8 & 27.2 & 33.5 & 24.7 & 83.9 & 36.7 & 78.8 & 58.7 & 30.5 & 84.8 & 38.5 & 44.5 & 1.7 & 31.6 & 32.4 & 45.5 \\
        DACS \cite{tranheden2021dacs} & 89.9 & 39.7 & 87.9 & 30.7 & 39.5 & 38.5 & 46.4 & 52.8 & 88.0 & 44.0 & 88.8 & 67.2 & 35.8 & 84.5 & 45.7 & 50.2 & 0.0 & 27.3 & 34.0 & 52.1 \\
        ProDA \cite{zhang2021prototypical} & 87.8 & 56.0 & 79.7 & 46.3 & 44.8 & 45.6 & 53.5 & 53.5 & 88.6 & 45.2 & 82.1 & 70.7 & 39.2 & 88.8 & 45.5 & 59.4 & 1.0 & 48.9 & 56.4 & 57.5 \\
        DAFormer \cite{tranheden2021dacs} & 95.7 & 70.2 & 89.4 & 53.5 & 48.1 & 49.6 & 55.8 & 59.4 & 89.9 & 47.9 & 92.5 & 72.2 & 44.7 & 92.3 & 74.5 & 78.2 & 65.1 & 55.9 & 61.8 & 68.3 \\
        HRDA \cite{hoyer2022hrda} & 97.12 & 78.03 & 90.83 & 60.49 & 47.42 & 57.63 & 59.73 & 70.38 & 91.17 & 44.02 & 93.43 & 79.04 & 54.46 & 94.62 & 85.26 & 85.01 & 67.97 & 63.16 & 68.10 & 73.05\\
        \rowcolor[gray]{0.85}HRDA + AFR & 96.97 & 77.11 & 90.87 & \textbf{63.06} & 50.28 & 58.98 & 63.87 & 71.39 & 91.39 & 46.62 & 94.09 & 78.15 & 53.04 & 94.40 & 83.17 & 85.70 & 74.07 & 60.62 & 68.63 & 73.81 {\scriptsize\textcolor{blue}{$\uparrow$0.76\%}}\\
        
        MIC \cite{hoyer2023mic} & 97.37 & \underline{80.37} & \underline{91.31} & 60.85 & 51.83 & 59.33 & 61.92 & 73.19 & \textbf{91.82} & 50.16 & 94.33 & \underline{80.17} & \underline{55.53} & 94.60 & \underline{86.04} & \underline{90.33} & 81.63 & \textbf{65.49} & 69.13 & 75.55\\

        \rowcolor[gray]{0.85}MIC+AFR (Ours) & \textbf{97.66} & \textbf{81.65} & \textbf{91.43} & \underline{61.99} & \underline{56.97} & \textbf{61.98} & \textbf{64.53} & \textbf{74.62} & 91.66 & 51.29 & 93.86 & \textbf{81.12} & \textbf{58.30} & \textbf{94.86} & 85.03 & \textbf{90.74} & \textbf{82.41} & 64.87 & \underline{70.33} & \textbf{76.60} {\scriptsize\textcolor{blue}{$\uparrow$1.05\%}} \\ 
        
        ERF \cite{song2025extended} & 97.01 & 78.16 & 91.31 & 60.18 & 56.61 & 59.22 & 63.62 & 68.41 & 91.67 & \textbf{52.95} & \underline{94.5} & 79.71 & 55.27 & \underline{94.65} & \textbf{86.1} & 90.14 & 81.55 & \underline{65.2} & \textbf{70.58} & 75.62 \\

        \rowcolor[gray]{0.85} ERF+AFR & \underline{97.38} & 80.11 & 91.16 & 60.6 & \textbf{58.28} & \underline{60.43} & \underline{64.5} & \underline{73.88} & \underline{91.75} & \underline{52.52} & \textbf{94.77} & 80.11 & 54.47 & 94.53 & 85.5 & 90.17 & \underline{81.76} & 64.48 & 70.23& \underline{76.14} {\scriptsize\textcolor{blue}{$\uparrow$0.52\%}}\\

        \bottomrule
    \end{tabular}
    }
\end{table*}

\begin{table*}[h]

    \centering
    \renewcommand{\arraystretch}{1.3} % Optional: Adjust row spacing
    \caption{Quantitative comparison of the adaptation from Synthia $\rightarrow$ Cityscapes dataset.}
    \label{syn}
    \resizebox{\textwidth}{!}{ % Scale table to page width
    \begin{tabular}{p{2.2cm}|*{20}{>{\centering\arraybackslash}p{0.75cm}}} % Adjust column widths
        \toprule
        Method & Road & S.walk & Build. & Wall & Fence & Pole & Tr.Light & Sign & Veget. & Terrain & Sky & Person & Rider & Car & Truck & Bus & Train & M.bike & Bike & mIoU \\ \hline
        
        ADVENT \cite{vu2019advent} & 85.6 & 42.2 & 79.7 & 8.7 & 0.4 & 25.9 & 5.4 & 8.1 & 80.4 & – & 84.1 & 57.9 & 23.8 & 73.3 & – & 36.4 & – & 14.2 & 33.0 & 41.2 \\
        DACS \cite{tranheden2021dacs} & 80.6 & 25.1 & 81.9 & 21.5 & 2.9 & 37.2 & 22.7 & 24.0 & 83.7 & – & 90.8 & 67.6 & 38.3 & 82.9 & – & 38.9 & – & 28.5 & 47.6 & 48.3\\
        ProDA \cite{zhang2021prototypical} & 87.8 & 45.7 & 84.6 & 37.1 & 0.6 & 44.0 & 54.6 & 37.0 & 88.1 & – & 84.4 & 74.2 & 24.3 & 88.2 & – & 51.1 & – & 40.5 & 45.6 & 55.5\\
        DAFormer \cite{tranheden2021dacs} & 84.5 & 40.7 & 88.4 & 41.5 & 6.5 & 50.0 & 55.0 & 54.6 & 86.0 & – & 89.8 & 73.2 & 48.2 & 87.2 & – & 53.2 & – & 53.9 & 61.7 & 60.9 \\
        HRDA \cite{hoyer2022hrda} & 84.89 & 45.73 & 89.13& 45.32 & 8.21 & 56.26 & 66.05 & 64.26 & 85.9 & -& 93.33 & 78.99 & 53.42 & 87.8 & - & 63.86 & -& 62.17 & 64.63 & 65.62\\
        
        \rowcolor[gray]{0.85}HRDA + AFR & 88.74& 49.77 & \textbf{89.37} & \textbf{49.05} & 2.87 & 58.26& 65.9 & 61.76 & \textbf{89.28} & - & 93.57 & 80.13 & 51.87 & 89.88 & - & \textbf{67.43} & - & 63.85 & 62.22 & 66.50 {\scriptsize\textcolor{blue}{$\uparrow$0.88\%}}\\
        
        MIC \cite{hoyer2023mic} &88.39 & 51.74 & 89.19 & 48.05 & 7.85 & 56.02 & 66.69& 62.09 & \underline{88.04} &- & 94.02 & 81.01 & 58.02 & \underline{90.84} &- & \underline{67.02} & -& 63.07 & \textbf{67.04} & 67.26\\
        
        \rowcolor[gray]{0.85}MIC+AFR (Ours) & 88.50 & 54.83 & 88.89 & \underline{48.42} & \textbf{8.74} & \textbf{59.33} & 66.43 & \textbf{63.67} & 87.41 & - & 93.86 & \textbf{81.77} & \textbf{59.22} & \textbf{91.18} & - & 66.76 & -& \textbf{68.36} & 65.44 & \textbf{68.30} {\scriptsize\textcolor{blue}{$\uparrow$1.04\%}} \\  
          ERF \cite{song2025extended} & \underline{89.67} & \underline{55.87} & 88.44 & 46.25 & 1.28 & 58.41 & \underline{66.76} & \underline{62.73} & 87.17 & - & \textbf{94.64} & \underline{81.5} & \underline{59.13} & 94.65 & - & 62.37 & - & 66.93 & 63.69 & 67.46 \\

          \rowcolor[gray]{0.85} ERF+AFR & \textbf{89.89} & \textbf{55.9} & \underline{89.23} & 47.24 & \underline{8.09} & \underline{59.29} & \textbf{67.62} & 62.52 & 87.73 & - & \underline{94.26} & 81.25 & 57.78 & 90.67 & - & 65.62 & -& \underline{67.67} & \underline{65.72} & \underline{68.16} {\scriptsize\textcolor{blue}{$\uparrow$0.70\%}} \\
        \bottomrule
        
    \end{tabular}
   
    }
\end{table*}
 %\vspace{-5pt}
\begin{table}[ht]
\centering
\caption{Runtime and memory consumption during training and inference on an RTX 4090.}
\label{run}
%\small
\resizebox{\linewidth}{!}{%
\begin{tabular}{l@{\hspace{10pt}}cc@{\hspace{10pt}}cc}
\toprule
\textbf{Method} & \multicolumn{2}{c}{\textbf{Training}} & \multicolumn{2}{c}{\textbf{Inference}} \\
 & Throughput & GPU Memory & Throughput & GPU Memory \\
\midrule
HRDA \cite{hoyer2022hrda} & 0.92 it/s & 20.47 GB &2.02 img/s & 9.45 GB \\
HRDA+AFR & 0.85 it/s & 20.47 GB & 1.88 img/s & 9.85 GB \\
MIC \cite{hoyer2023mic} & 0.73 it/s & 20.58 GB & 1.92 img/s & 9.91 GB \\
MIC+AFR & 0.70 it/s & 20.51 GB &1.91 img/s & 9.91 GB \\
ERF \cite{song2025extended} & 0.75 it/s & 18.78 GB & 2.24 img/s & 8.21 GB \\
ERF+AFR & 0.79 it/s & 18.83 GB & 2.21 img/s & 8.21 GB \\
\bottomrule
\end{tabular}
}
%\vspace{-20pt}
\end{table}
Table \ref{gt} shows that our proposed AFRDA achieves 76.60$\%$ mIoU, outperforming the baseline MIC by +1.05 mIoU. AFRDA demonstrates leading performance in 11 out of 19 classes for the GTA V$\rightarrow$ Cityscapes adaptation task, including small and rare classes such as ``Fence'', ``Pole'', ``Traffic Light'', ``Traffic Sign'', and ``Train'', which are typically difficult to predict. Moreover, the core component AFR of the AFRDA framework exhibits promising performance when integrated into the HRDA framework, leading to a +0.76 mIoU improvement. Later, we present the quantitative results for the Synthia $\rightarrow$ Cityscapes adaptation in Table \ref{syn}. As shown in Table \ref{syn}, AFRDA outperforms all state-of-the-art (SOTA) methods and surpasses the baseline MIC by +1.04 mIoU on the Synthia $\rightarrow$ Cityscapes adaptation task. Moreover, AFRDA demonstrates its effectiveness in predicting challenging classes in this setup as well. Furthermore, integrating AFR with the HRDA module results in an improvement of +0.88 mIoU over the HRDA framework, demonstrating the feasibility of AFR as a plug-and-play module. We also plugged our AFR module on top of a very recent UDA method named ERF \cite{song2025extended} and showed that our AFR module improves the overall result for both GTA V$\rightarrow$ Cityscapes and Synthia $\rightarrow$ Cityscapes adaptation tasks.  We show the qualitative results of our proposed AFRDA in Fig. \ref{cityp} and Fig. \ref{meshp} for both the city and forest environment adaptation tasks, respectively. As shown in Fig. \ref{cityp}, our model performs better than other SOTA methods in GTA V $\rightarrow$ Cityscapes and Synthia $\rightarrow$ Cityscapes adaptation. The results show that AFRDA achieves superior segmentation of ``Wall," ``Fence," ``Vegetation," and ``Sidewalk," producing sharper boundaries, smoother edges, and more accurate predictions of minor categories such as ``Traffic Signs" and ``Poles." The reason is that the AFR module explicitly leverages boundary details and enhances contextual awareness, which improves the model’s ability to distinguish sidewalks from roads and fences from walls while accurately predicting poles and traffic signs, leading to more robust and precise segmentation in complex urban scenes. Fig. \ref{meshp} demonstrates the superiority of our proposed AFRDA framework in a forest environment. While all other methods struggle to accurately predict grass, especially when it appears drier or yellowish, our model successfully predicts ``Grass", ``Bushes", ``Sky", and other elements with high precision. %We also report the throughput and GPU memory usage of various UDA methods with and without AFR in Table~\ref{run}. Adding AFR to existing models results in minimal training slowdown (e.g., only 7.6$\%$ for HRDA) and maintains comparable or improved inference speed. Notably, AFR slightly reduces training GPU memory in MIC (20.58$\rightarrow$20.51 GB), likely due to its structured attention reducing intermediate feature redundancy. AFR also preserves or enhances training speed when added to ERF, highlighting its lightweight and efficient design. Overall, AFR delivers accuracy gains with negligible computational cost.
We also report the throughput and GPU memory usage of various UDA methods with and without AFR in Table~\ref{run}. Adding AFR to existing models results in minimal training slowdown (e.g., only 7.6$\%$ for HRDA) while maintaining comparable or improved inference speed. Notably, AFR slightly reduces training GPU memory in MIC (20.58$\rightarrow$20.51~GB), likely due to structured attention reducing intermediate feature redundancy. It also preserves or enhances training speed when added to ERF, highlighting its lightweight, efficient design. Overall, AFR delivers accuracy gains with negligible computational cost.
 
%\vspace{-5pt}
\subsection{Ablation Studies}
\begin{figure*}[t] \vspace{-10pt}
    \centering
    \includegraphics[width=1\linewidth]{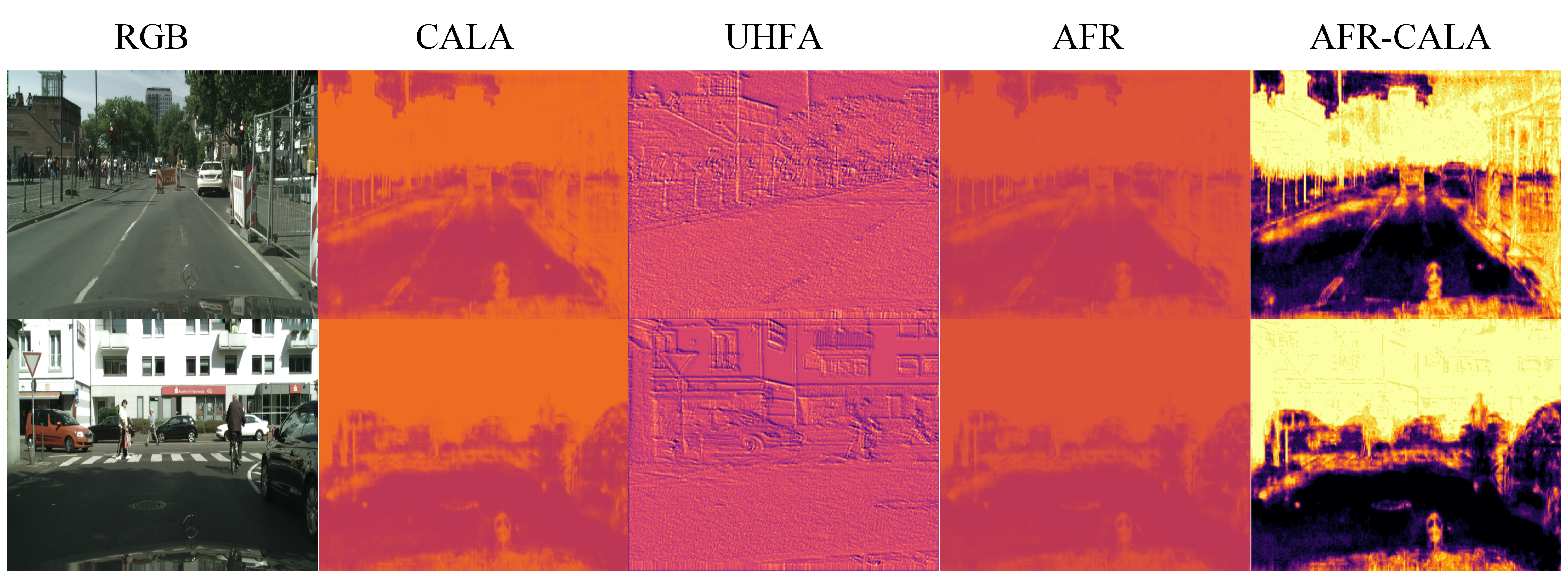} % Replace 'example-image' with your immixedage filename
    \caption{\small Visualization of the attention maps from the AFR module. }
    
    \label{at}
\vspace{-15pt}
\end{figure*}
We evaluate the effectiveness and design of our proposed AFR module by conducting ablation studies for the GTA V $\rightarrow$ Cityscapes adaptation task, as shown in Table \ref{ab}.
%To assess the effectiveness of the Attentive Feature Refinement (AFR) module, we conduct ablation studies by selectively disabling key components and analyzing their impact on segmentation performance. The experiments focus on evaluating the contributions of individual modules, the role of uncertainty estimation, and the significance of boundary refinement. The results from these experiments are discussed in the following subsections.
\begin{table}[ht]
    \caption{Ablation study results showing the impact of different components on mIoU performance.}
    \label{ab}
    \centering
    \renewcommand{\arraystretch}{1.1}
    \begin{tabular}{p{5.5cm} |c| c}
        \hline
        \multirow{2}{*}{Model Variation} & \multicolumn{2}{c}{mIoU (\%)} \\  
        \cline{2-3}
        & Absolute & $\delta_{\text{AFR}}$ \\  
        \hline
        Baseline (No AFR) & 75.55 & -1.05 \\  
        \hline
        AFR w/o CALA & 76.04 & -0.56 \\  
        AFR w/o UHFA & 75.86 & -0.74 \\  
        \hline
        AFR w/o HR Uncertainty (CALA) & 75.17 & -1.43 \\  
        AFR w/o LR Logits Uncertainty (UHFA) & 76.00 & -0.60 \\  
        \hline
        AFR w/o Boundary (CALA \& UHFA) & 75.58 & -1.02 \\  
        AFR w/o Boundary in CALA & 75.20 & -1.4 \\  
        AFR w/o Boundary in UHFA & 75.65 & -0.95 \\  
        \hline
        Full AFR & 76.60 & 0\\  
        \hline
    \end{tabular}
%\vspace{-15pt}
\end{table}

\subsubsection{\textbf{Visualization of AFR module}}
To demonstrate the interpretability and effect of our AFR module, we visualize key attention components in Fig. \ref{at}. The CALA attention map highlights large and confident semantic regions such as the road, sky, and building facades with high attention values, consistent with its role in providing semantic-aware global guidance. While some smaller objects like poles and traffic signs are visible, they receive relatively smooth and low-detail attention, indicating limited sensitivity to fine structures. In contrast, the UHFA attention map, shown from one high-resolution crop for clarity, focuses on high-frequency spatial signals and enhances edges and boundaries, particularly around poles, buildings, traffic signs, and pedestrians. The final attention map integrates both sources through a learnable fusion. To visualize the specific effect of UHFA, we show the difference between the final attention and CALA, which reveals strong activations around structural boundaries, class transitions, and object contours. In both examples, outlines of poles, traffic signs, and building boundaries are distinctly visible. These higher values confirm UHFA’s role in boosting attention in boundary-sensitive regions while preserving the broader semantic structure from CALA. Although we do not provide a binary UHFA mask due to threshold variability, the visualizations show that UHFA suppresses flat regions and retains high-frequency boundary content, supporting our claim that AFR enhances high-resolution feature refinement along class boundaries and object edges.
\subsubsection{\textbf{Effect of CALA and UHFA}}
%We conduct experiments where each module is individually removed to evaluate the contributions of the Class-Aware Logits-Based Attention (CALA) and the Uncertainty-Suppressed HR Feature Attention (UHFA) modules. The results indicate that disabling CALA leads to a minor drop in performance, achieving a mIoU of 76.04$\%$, whereas disabling UHFA results in a slightly greater reduction in accuracy, achieving a mIoU of 75.86$\%$. These findings indicate that both modules contribute to AFR’s effectiveness, with UHFA having a slightly stronger impact. While each module independently improves segmentation beyond the baseline, their combination yields the most significant gains. This reinforces that AFR benefits from the complementary interaction between CALA and UHFA, integrating class-aware refinement with uncertainty-suppressed high-resolution features to enhance both local and global representations.
We conduct experiments where each module is individually removed to evaluate the contributions of the Class-Aware Logits-Based Attention (CALA) and the Uncertainty-Suppressed HR Feature Attention (UHFA) modules. The results show that disabling CALA leads to a minor drop in performance (76.04$\%$ mIoU), whereas disabling UHFA results in a slightly greater reduction (75.86$\%$ mIoU). These findings indicate that both modules contribute to AFR’s effectiveness, with UHFA having a slightly stronger impact. While each module independently improves segmentation beyond the baseline, their combination yields the most significant gains, reinforcing that AFR benefits from the complementary interaction between CALA and UHFA to enhance both local and global representations.

%These findings suggest that both modules contribute to AFR’s overall effectiveness, with UHFA playing a slightly more dominant role. Although each module independently enhances segmentation performance beyond the baseline, the full AFR model demonstrates that their combination provides the most significant improvement. The results further reinforce the notion that AFR does not rely on a single mechanism but instead benefits from the complementary nature of CALA and UHFA. Their integration balances class-aware refinement and uncertainty-suppressed high-resolution feature enhancement, optimizing local and global feature representations for improved segmentation accuracy.

\subsubsection{\textbf{Impact of Uncertainty Estimation}}
%\vspace{-3pt}
We conduct experiments where uncertainty estimation is removed separately from the high-resolution features in CALA and the low-resolution logits in UHFA. The results show that removing high-resolution uncertainty leads to the largest observed performance drop (75.17$\%$ mIoU), while removing low-resolution logit uncertainty achieves a mIoU of 76$\%$. This indicates that uncertainty suppression at high resolution is critical for AFR’s effectiveness. Since high-resolution uncertainty is multiplied with low-resolution logits, its removal disrupts confidence-based refinement, reducing the effectiveness of global class priors and increasing spatial inconsistencies. Although high-resolution features are still processed through attention and high-frequency extraction, they lose proper weighted low-resolution guidance. Unlike the baseline, which maintains stable feature flow, AFR without high-resolution uncertainty misaligns global context with local details, leading to segmentation errors that other components cannot compensate for.

\begin{figure*}[h]
    \centering
    \includegraphics[width=0.95\linewidth]{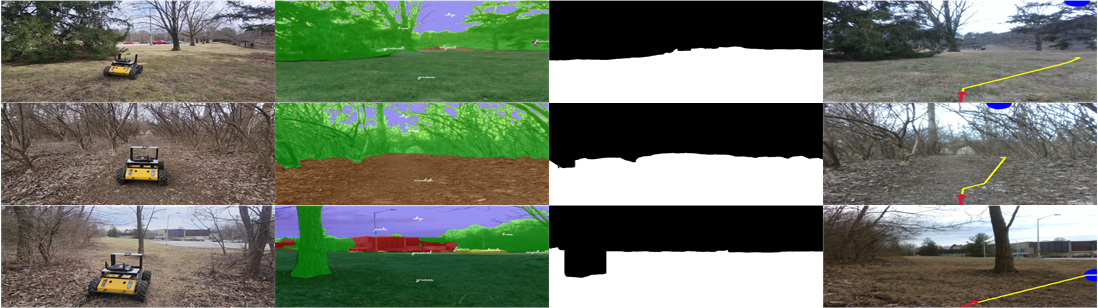}
    \caption{Navigation behavior in the forest environment. From left to right, we show the third-person view, segmentation result, navigable image, and planning result.} 
    \label{fig:ro} 
    \vspace{-15pt}
\end{figure*}
\subsubsection{\textbf{Significance of Boundary Refinement}}
%We also conduct ablation studies by removing high-frequency components from CALA and UHFA separately and together. Removing boundary refinement only from CALA leads to a greater performance drop (75.20\%) than removing it from both modules (75.58\%), while removing it only from UHFA results in 75.65\%. This suggests that class-aware boundary refinement ensures proper alignment of class priors with spatial details, directly influencing how low-resolution logits interact with high-resolution features. Without it, the model relies on misaligned boundary cues, increasing segmentation errors. Removing boundary refinement from both modules reduces this misalignment as the model adapts using other available features like uncertainty weighting and contextual priors. With boundary refinement in both CALA and UHFA, segmentation improves as class-prior boundaries align with spatially consistent boundaries. CALA refines low-resolution logits, while UHFA sharpens object edges and reduces noise; demonstrating the effective design of AFR. 
We also conduct ablation studies by removing high-frequency components from CALA and UHFA separately and together. Removing boundary refinement only from CALA leads to a greater performance drop (75.20\%) than removing it from both modules (75.58\%), while removing it only from UHFA results in 75.65\%. This suggests that class-aware boundary refinement ensures proper alignment of class priors with spatial details, directly influencing how low-resolution logits interact with high-resolution features. Without it, the model relies on misaligned boundary cues, increasing segmentation errors. 
\begin{table}[h]
\centering
%\scriptsize
\setlength{\tabcolsep}{2pt}
\caption{IoU comparison for small structure classes across different model variants.} \label{hf}
\begin{tabular}{@{}lcccccc@{}}
\toprule
\textbf{Method} & Pole & Tr. Light & Tr. Sign & Rider & M.bike & Bike \\
\midrule
Full AFR       & 61.98 & 64.53 & 74.62 & 58.30 & 64.87 & 70.33 \\
w/o HF Cues    & 60.84 & 63.14 & 72.39 & 56.25 & 64.49 & 70.43 \\
No AFR         & 59.33 & 61.92 & 73.19 & 55.53 & 65.49 & 69.13 \\
\bottomrule
\end{tabular}
\end{table}
Removing boundary refinement from both modules reduces this misalignment as the model adapts using other available features like uncertainty weighting and contextual priors. With boundary refinement in both CALA and UHFA, segmentation improves as class-prior boundaries align with spatially consistent boundaries; CALA refines low-resolution logits, while UHFA sharpens object edges and reduces noise, demonstrating the effective design of AFR. To further validate the contribution of high-frequency cues to small object segmentation, we present class-wise IoU results in Table \ref{hf}. Removing high-frequency information consistently degrades performance across all small structure categories. For example, compared to the full AFR model, the IoU drops by 1.14$\%$ for Pole, 1.35$\%$ for Traffic Light, and 1.94$\%$ for Rider. These results confirm that UHFA’s high-frequency refinement improves detection of small and thin classes by enhancing edge sensitivity and preserving fine-grained structural details.

%Together, they eliminate misalignment and improve accuracy.

%demonstrating that AFR effectively integrates all components.
%\vspace{-5pt}

\subsection{Navigation Missions}
\begin{comment}
We tested our proposed AFRDA model, trained with the RUGD $\rightarrow$ MESH setup, by combining it with the POVNav planner \cite{pushp2023povnav} to see how well it works in real-world navigation. We ran our navigation system in the forest environment in front of our lab (shown in Fig. \ref{fig:ro}). 
\begin{figure} [h]
   \centering
    
    \includegraphics[width=1\linewidth]{robot.png}
    \caption{Navigation behavior in the forest environment. We show the third-person view, segmentation result, navigable image result, and the planning result from the left column to the right column, respectively.} \label{fig:ro} 
    \vspace{-15pt}
\end{figure}
During navigation, we used images with a resolution of 640×480 pixels. The system was deployed on a Husky robot equipped with an NVIDIA GeForce RTX 2060 GPU. Our AFRDA model took 0.72 seconds to complete the segmentation task. Since other processing steps were also needed for perception, the entire process took about 0.77 seconds. We set the robot's speed to 0.1 m/s and tested it on paths of around 10 meters in length. Although the robot moved slowly, our method successfully avoided non-navigable areas and reached the goal. 
%This shows that our AFRDA model is effective for navigation in unstructured environments.
%\vspace{-3pt}
\end{comment}

We evaluated our AFRDA model, trained on the RUGD $\rightarrow$ MESH setup, by integrating it with the POVNav planner \cite{pushp2023povnav} and deploying it on a Husky robot in a forest near our lab (Fig. \ref{fig:ro}). Using 640×480 images on an RTX 2060 GPU, AFRDA took 0.72 s for segmentation, and the full pipeline took ~0.77 s. The robot moved at 0.1 m/s over ~10 m paths, successfully avoiding non-navigable areas to reach the goal.

\section{CONCLUSIONS}
We propose a model named AFRDA for improving unsupervised domain-adaptive semantic segmentation. Our method is based on a self-training framework that explores both local and global details along with the boundary information. To achieve this, we developed a module named Attentive Feature Refinement that refines the high-resolution features based on the low-resolution logits. The AFR leverages the semantic awareness of low-resolution logits and the spatial awareness of high-resolution features to improve semantic segmentation.  We tested our method on benchmark datasets, and our model outperforms the SOTA by a significant margin. Moreover, we deployed our framework in a robotic vehicle to navigate in unstructured environments.


%%%%%%%%%%%%%%%%%%%%%%%%%%%%%%%%%%%%%%%%%%%%%%%%%%%%%%%%%%%%%%%%%%%%%%%%%%%%%%%%

%%%%%%%%%%%%%%%%%%%%%%%%%%%%%%%%%%%%%%%%%%%%%%%%%%%%%%%%%%%%%%%%%%%%%%%%%%%%%%%%

%%%%%%%%%%%%%%%%%%%%%%%%%%%%%%%%%%%%%%%%%%%%%%%%%%%%%%%%%%%%%%%%%%%%%%%%%%%%%%%%

%\section*{ACKNOWLEDGMENT}

\bibliographystyle{IEEEtran}
\bibliography{ref}

\end{document}